# Rethinking Deep Learning:
# Non-backpropagation and Non-optimization Machine Learning Approach Using Hebbian Neural Networks

Kei Itoh*

November 7, 2024

## Abstract

Developing strong AI could provide a powerful tool for addressing social and scientific challenges. Neural networks (NNs), inspired by biological systems, have the potential to achieve this. However, weight optimization techniques using error backpropagation are not observed in biological systems, raising doubts about current NNs approaches. In this context, Itoh (2024) solved the MNIST classification problem without using objective functions or backpropagation. However, weight updates were not used, so it does not qualify as machine learning AI. In this study, I develop a machine learning method that mimics biological neural systems by implementing Hebbian learning in NNs without backpropagation and optimization method to solve the MNIST classification problem and analyze its output. Development proceeded in three stages. In the first stage, I applied the Hebbian learning rule to the MNIST character recognition algorithm by Itoh (2024), resulting in lower accuracy than non-Hebbian NNs, highlighting the limitations of conventional training procedures for Hebbian learning. In the second stage, I examined the properties of individually trained NNs using norm-based cognition, showing that NNs trained on a specific label respond powerfully to that label. In the third stage, I created an MNIST character recognition program using vector norm magnitude as the criterion, achieving an accuracy of approximately 75%. This demonstrates that the Hebbian learning NNs can recognize handwritten characters without objective functions, backpropagation, optimization processes, and large data set. Based on these results, developing a mechanism based on norm-based cognition as a fundamental unit and then increasing complexity to achieve indirect similarity cognition should help mimic biological neural systems and contribute to realizing strong AI.

* Ehime University Graduate School of Science and Technology. The author conducted this research independently while affiliated with Ehime University, and utilized the university's library and electronic resources. The research was conducted without specific financial support from external funding sources. The author declares no conflict of interest.

## 1. Introduction

The development of strong AI and the accompanying arrival at the technological singularity could be a powerful tool for addressing various issues in society and science [e.g., 1]. Given that neural networks [e.g., 2,3] and deep learning, which are structured to mimic biological neural systems, hold the potential to create strong AI [e.g., 4]. NNs-based AI has been implemented with practical applications through error backpropagation and objective function-based weight optimization methods [e.g., 5]. However, these methods are not observed in biological neural systems [e.g., 6,7,8], raising significant doubts as to whether strong AI can be achieved through the current NNs-AI implementation theory.

In this context, Itoh (2024) [8] demonstrated the fundamental information propagation capability of NNs by solving the Modified National Institute of Standards and Technology database (MNIST) [9] classification problem under non-weight-update conditions, aiming to emulate biological neural systems and develop strong AI without using error backpropagation or objective functions. This study solved the handwritten character recognition problem with an accuracy of approximately 80% by comparing output vectors utilizing the concept of distributed representations [10].

Since Itoh (2024) solved the classification problem without weight updates, these NNs and algorithms cannot be considered machine learning AI. Biological organisms with nervous systems, including humans, learn and remember based on Hebbian learning principles [7]. Thus, incorporating machine learning methods is essential when using NNs to mimic nervous systems. Hebbian learning is a rule generally stated as "synapses between two neurons that fire together are strengthened", forming the fundamental mechanism for learning, memory, and recognition in biological systems. As previously mentioned, current NNs-AI alters weights (synapses in the context of nervous systems) via optimization by introducing objective functions rather than using Hebbian learning principles. Unlike Hebbian learning, optimization methods such as error backpropagation are highly divergent from the memory and learning mechanisms observed in nervous systems.

This research aims to develop a machine learning method that emulates biological neural systems by implementing Hebbian learning principles in NNs based on Itoh's (2024) design philosophy, solving the MNIST classification problem, and analyzing its output. The development of this machine learning method will proceed in stages. The first stage directly applies Itoh's (2024) MNIST classification solution to Hebbian learning NNs and calculates accuracy. In the second stage, NNs trained individually on each MNIST digit label will be used, and the applicability of Hebbian learning to NNs will be discussed by comparing the norms of the output vectors of these NNs. The third stage extends the method from the second stage to solve the MNIST classification problem by comparing the norms of the output vectors from individually trained NNs. Furthermore, based on these developments and results, a discussion on recognition using vector norms in NNs and biological neural systems will be conducted.

## 2. Calculation Method

This section describes the calculation method standard for all NNs-AI used in this study. Specific methods and workflows employed individually will be described later.

The neural network in this study is constructed in Python (version 3.10.11) using the PyTorch library (version 2.2.1). The dimensions of the input, hidden, and output layers are all set to 784, and all layers of the NNs are fully connected, using the ReLU function as the activation function. While Itoh (2024) used the step function as the activation function, using it in this study did not yield good results in Sections 4 and 5. This may be because the ReLU function, as a function form, allows for more diverse representations than the step function. He initialization [11] is used for weight initialization in the NNs.

In this study, all weight update rules in the NNs follow Hebbian learning principles. To implement Hebbian learning in the NNs, the weight update matrix between hidden layers $i$ and $i+1$, denoted as $\mathbf{W}_i$, is given by the following formula when data is input:

$$\mathbf{W}_i = \eta \mathbf{a}_i \mathbf{a}_{i+1}^{\mathrm{T}}$$

Where $\eta$ is the learning rate, $\mathbf{a}_i$ is the activation state matrix of hidden layer $i$, and $\mathbf{a}_{i+1}^{\mathrm{T}}$ is the transposed activation state matrix of hidden layer $i+1$. The matrix product of these activation state matrices determines whether both neurons at each weight's ends are active. The weight update amount is then calculated by multiplying by the learning rate. Additionally, weight correction updates are performed to model the decrease in synaptic strength over time to prevent weights from growing indefinitely in strength. The total weight update $\Delta\mathbf{W}_i$ is divided by the number of weight elements (since the layers are fully connected, 784*784=614656) and this amount is subtracted from all weights to keep the overall weight constant.

However, as exemplified by spike-timing-dependent plasticity [e.g., 12], learning rules in actual nervous systems are far more complex. In this study, a simplified weight update rule is adopted solely to verify the effect of Hebbian learning on the NNs; more advanced learning rules would likely be necessary to implement higher-level intelligence.

## 3. MNIST Image Recognition Using Distributed Representation Average Vectors with Hebbian Learning

### 3.1 Calculation Conditions

First, I apply the Hebbian learning rule to the NNs and MNIST solution algorithm Itoh (2024) constructed to obtain accuracy rates. The primary calculation conditions and workflow refer to Itoh (2024), although I have made some additions and modifications, as described here. Specifically, the training data is split in half: one part is used for Hebbian learning, and the other is used for average vector calculations. The learning process is added and defined during network construction. While Itoh (2024) used Euclidean distance to calculate vector similarity, this study also incorporates cosine similarity.

### 3.2 Results

Figure 1 shows the results of MNIST accuracy rates using distributed representation average vector comparison with Hebbian learning-implemented NNs—figure 1. A displays the accuracy rates for the MNIST classification problem when the learning rate is fixed, and the number of hidden layers is varied—figure 1. B shows the accuracy rates when the number of hidden layers is fixed and the learning rate is varied.

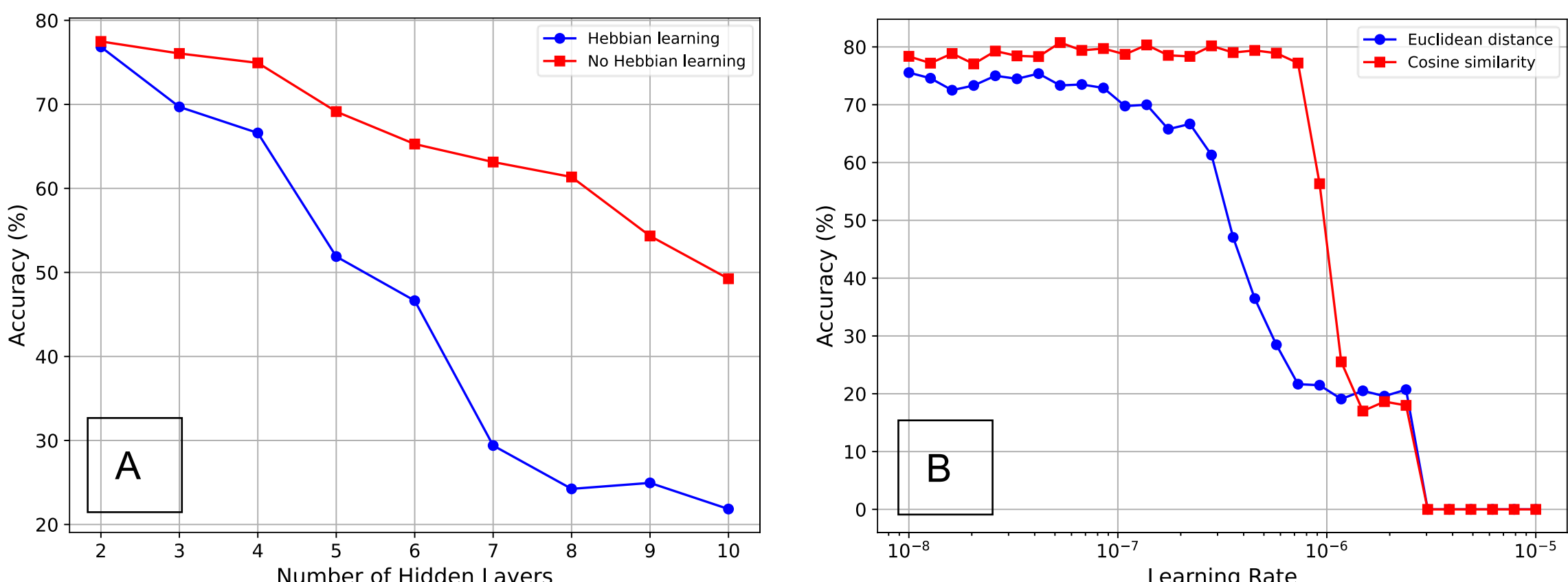

**Figure 1.** MNIST accuracy rates using distributed representation average vectors with Hebbian learning-implemented NNs. The learning rate is set to $10^{-7}$ in A. The network has three hidden layers in B, and the horizontal axis is on a logarithmic scale.

Figure 1. A shows that implementing the Hebbian learning rule does not improve accuracy. On the contrary, NNs without Hebbian learning achieve higher accuracy when there are three or more hidden layers. This suggests that conventional training procedures may not suit the Hebbian learning rule well. Figure 1. B shows that accuracy sharply declines when the learning rate becomes relatively large. This is confirmed to be due to the rapid increase in the absolute values of the output vector components, eventually diverging to infinity (causing output values that are too large to be processed).

### 3.3 Inapplicability of the Hebbian Learning Rule to Discrete Information

The Hebbian learning rule is a capability that organisms acquire through evolution. Organisms consistently receive continuous information aligned with the arrow of time, which is input into their nervous systems [e.g., 6,7,12]. However, conventional input information for NNs, including that used by Itoh (2024), is discrete and lacks a temporal dimension. Suppose the Hebbian learning rule only applies to continuous information and is inapplicable to discrete details. In that case, the observed lack of improvement—or even decrease—in accuracy when simply using the Hebbian learning rule in conventional training procedures can be explained. In other words, to make NNs with Hebbian learning practical, it is necessary to alter conventional training procedures significantly.

## 4. Norm Comparison Using Individually Trained NNs

### 4.1 Introduction to Program Design

In the development of deep learning AI, each layer of an NNs is typically treated as a vector, as is the output data, allowing the definition of distance and similarity within vector space, thereby imbuing data with meaning (distributed representation). However, biological neural systems likely lack mechanisms for perceiving or calculating the vector similarity of the information they convey. (This assertion is my hypothesis and has not been widely studied or advocated previously. A more detailed discussion will be provided in Section 6.) Instead, I believe that behaviors observed in biological systems should be understood based primarily on the magnitude of vector norms rather than vector similarity. For example, reflex actions—one of the most primitive behaviors in organisms—are unlikely to be based on vector similarity. Reflexes involve binary decisions, such as whether or not to perform a specific action, which are based on the magnitude of input norms or processed internal information. Regarding actual living creatures, the sea slug (Aplysia) performs reflexive gill contractions in response to stimulation of the body surface [13]. In the nervous system of Aplysia, the information from this stimulus is not processed as a distributed representation. Instead, it undergoes a simple binary process to determine whether there was intense neural firing.

In this section, I aim to develop a new machine learning method based on NNs, considering the judgment mechanism based on the non-distributive norm of such biological neural systems and the inapplicability of Hebbian learning rule to discrete information observed in Section 3. The MNIST dataset will continue to be used. Still, to introduce continuity in the input information, I attempt a method in which only a specific label is input into the NNs for Hebbian learning (this process will hereafter be referred to as "individual learning"). Individual learning is applied to all labels, and the characteristics of these individually trained NNs are examined by comparing the magnitudes of the vector norms concerning the test data. Specifically, when data from a particular label is input, I compare it with the more significant norm: the NNs individually trained on that label or other NNs.

### 4.2 Calculation Conditions and Workflow

The structure of the NNs and the Hebbian learning rule are the same as in the previous sections. Based on this structure, I modify the training procedure and verification method from conventional approaches. First, I create an individually trained NNs for each of the 0–9 labels using only the training data for a specific label. Next, I input the test data with the same label into this individually trained NNs and obtain L2 norm (all of the following norms are calculated using the L2 norm) of the output vector. The same test data is then input into all other NNs, and their norms are also obtained. I compare these norms to determine which is more extensive.

Additionally, for verification, I compare the norm magnitudes of NNs trained on all MNIST data, NNs trained on a uniform subset of 1/10 of the data for each label, and untrained NNs. To standardize the number of data points input during Hebbian learning, I adjusted the number of data points used in training so that all NNs would be uniform. I set the number to match the number of data points in the MNIST label with the fewest examples (label 5, with 5,420 data points for training). I also calculate the results when the number of training data points was limited to 1, 10, and 100 for each label. Furthermore, I apply norm normalization to the MNIST data.

The simplified workflow for the program in this section is as follows:

1. Library Import:

Import the necessary Python libraries (torch, torchvision, numpy).

2. Loading and Preprocessing MNIST Data:

Scale the MNIST pixel values to a range of 0–1. Load both the training and test data, separating each by label. Limit the number of training data points for each label. Normalize the norm of all data.

3. Defining Network Structure:

Define a fully connected, feedforward structure and set ReLU as the activation function. Define the learning process based on the Hebbian learning rule.

4. Training NNs:

Deploy the individually trained NNs for each label and other NNs to the GPU and train each with the appropriate training data. The seed value is set, and the weights are initialized to the same value for all models.

5. Vector Norm Magnitude Comparison:

Input the test data for a specific label into the NNs individually trained on that label and obtain the norm of the output vector. Input the same data into the other NNs, obtain their output vector norms, and define a function to compare these magnitudes. This function is used to compare the norms for each individually trained NNs.

6. Saving the results:

Saving the various necessary information.

**4.3 Results**

Figure 2 compares the magnitude of the output vector norms when test data of a specific label is input into the NNs individually trained on that label versus other NNs. The count increases when the norm of the NNs individually trained on a specific label is larger.

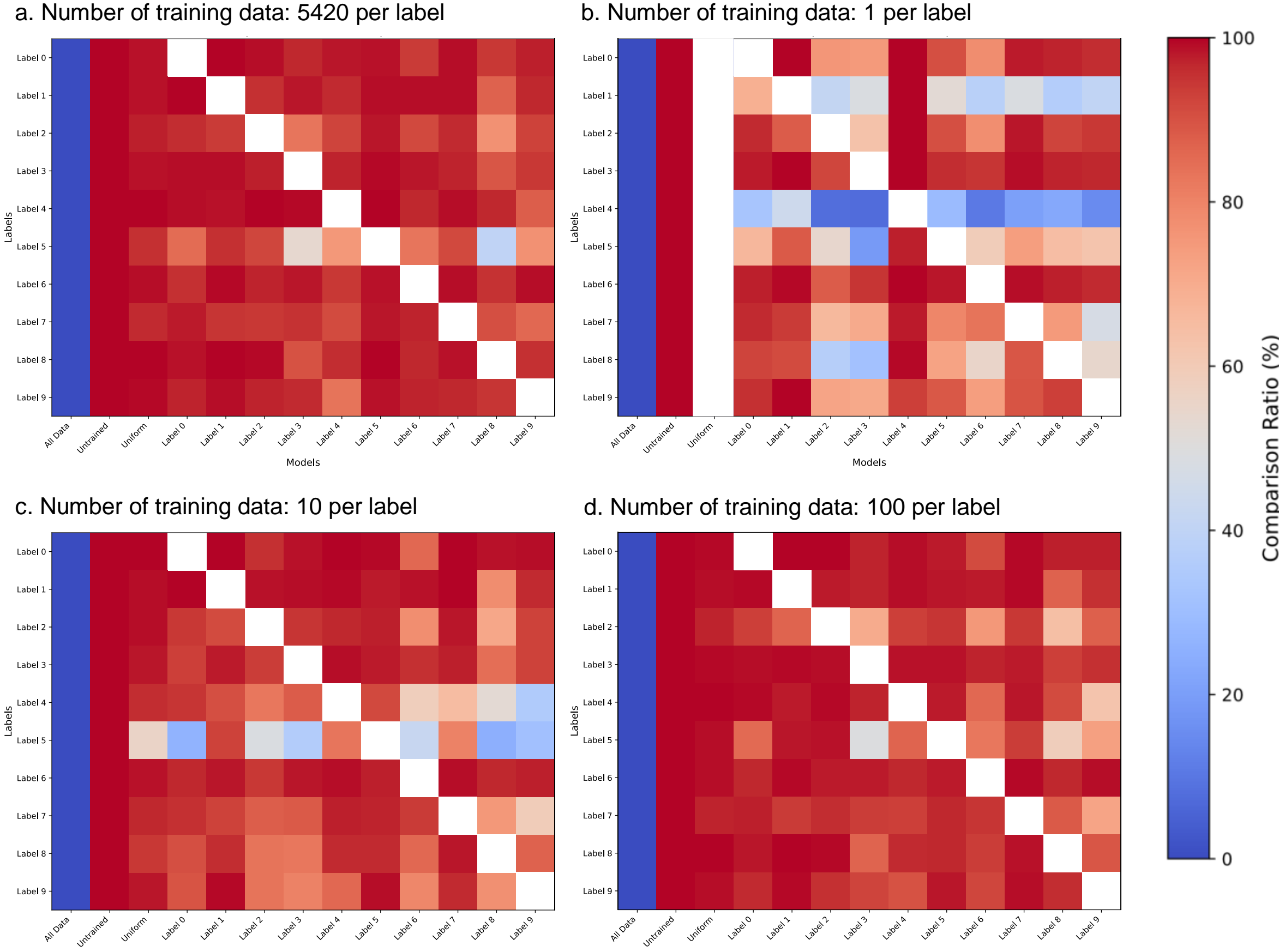

**Figure 2.** Comparison of output vector norms for individually trained NNs. These heatmaps show the results of comparing the output norms of label-specific individually trained NNs for each test data label with those of other models. The vertical axes represent the labels in the test data, the horizontal axes show the models being compared, and the color intensity in the heatmaps indicates the percentage of cases where the output norm of the label-specific individually trained NN for the input label exceeds that of the other models. The learning rate is set to $10^{-7}$, and the networks have three hidden layers. "All data" refers to the comparison with NNs trained on the full MNIST dataset, "Untrained" refers to the comparison with untrained NNs, and "Uniform" indicates the comparison with NNs trained on a uniform 1/10 portion of the MNIST dataset for each label.

Figure 2a shows that NNs individually trained on a specific label has a more significant output vector norm when tested with that label data than NNs trained on other labels. This result is generally consistent across labels except for label 5. When label 5 data are input, the NNs individually trained on labels 3 and 8 often exhibit more significant output vector norms than those trained on label 5. This likely occurs because the shape of the digit 5 resembles that of digits 3 and 8, causing the NNs trained on labels 3 and 8 to respond similarly or even more strongly to label 5 data. The NNs trained on all data have a more significant output vector norm than all individually trained NNs. At the same time, the untrained NNs exhibit more minor output vector norms than all individually trained NNs.

These findings suggest a proportional relationship between output vector norms and the amount of training data. The Uniform NNs shows more minor output vector norms than all individually trained NNs, indicating that individually trained NNs exhibit selective responsiveness to data of their specific label.

Figure 2b demonstrates that even when the number of training data per individually trained NN is limited to just one, the NN specifically trained on the input label tends to respond more strongly to the corresponding input than other NNs in most comparisons. Figures 2c and 2d show that as the number of training data increases to 10 and 100, the NNs acquire stronger selectivity. Furthermore, with 100 training samples, the results are comparable to those obtained with 5420 samples. These findings suggest that Hebbian learning-based NNs do not require large datasets and can effectively learn from a small number of samples. This characteristic aligns with the learning processes of biological systems, which do not rely on extensive datasets, making Hebbian learning-based NNs more biologically plausible than traditional NN learning processes.

These results show that individually trained NNs respond selectively to their respective labels. When Hebbian learning is applied to data of a specific label, the connections between neurons are strengthened, resulting in a more robust response to that label. This phenomenon aligns with the memory formation mechanism through synaptic plasticity, a well-established concept in neuroscience [14]. For example, in long-term potentiation (LTP), synaptic connections between neurons that receive repeated stimulation are strengthened, increasing selective responses to specific stimuli [15]. Thus, applying the Hebbian learning rule to NNs in the manner described in this study could potentially serve as a way to mimic biological neural systems.

## 5. MNIST Image Recognition Using Individually Trained Models and Norm Comparison

### 5.1 Calculation Conditions and Workflow

Using the selective responsiveness of individually trained NNs to specific labels demonstrated in Section 4, I developed an NNs program capable of MNIST image recognition. First, using the training data, I create individually trained NNs for each of the 0–9 labels. Next, I input test data into each of these NNs, compare the magnitudes of their output vector norms, and assign the label of the NNs with the most prominent norm as the predicted value. I then calculate the predicted values for all test data and compute the overall accuracy. As in Section 4, corrections are made by limiting the number of data points and applying norm normalization.

The simplified workflow for the program in this section is as follows:

1. Library Import:

Import the necessary Python libraries (torch, torchvision, numpy, etc.).

2. Loading and Preprocessing MNIST Data:

Scale the MNIST pixel values to a range of 0–1. Load both the training and test data, organizing each by label. Limit the number of training data points for each label and normalize the norm of all data.

3. Defining Network Structure:

Define a fully connected, feedforward structure and set ReLU as the activation function. Define the learning process based on the Hebbian learning rule.

4. Training NNs:

Deploy the individually trained NNs for each label to the GPU and train each NNs with its corresponding training data. The seed value is set, and the weights are initialized to the same value for all models.

5. Accuracy Calculation Using Norm Magnitude Comparison:

Define a function that inputs test data into each individually trained NNs, obtains the norm of the output vector, and compares the magnitudes. The label of the NNs with the most prominent norm is used as the predicted value, and accuracy is then calculated.

6. Saving Results:

Save various required information.

## 5.2 Results

Figure 3 shows the accuracy of the MNIST recognition task using individually trained NNs, with hidden layers ranging from 2 to 15 and learning rates from $10^{-1}$ to $10^{-9}$. Figure 4 displays the accuracy for each label.

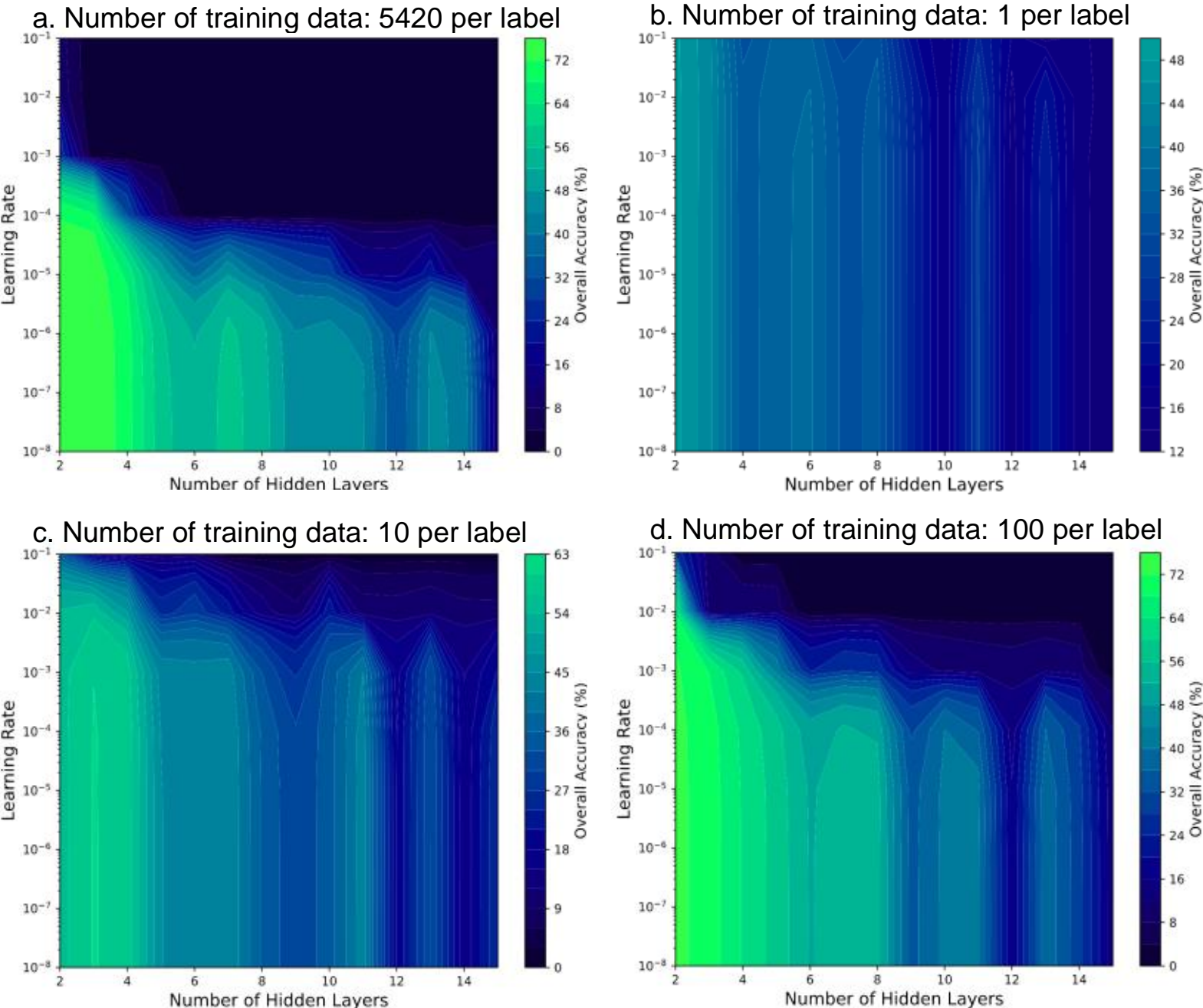

**Figure 3.** Accuracy of the MNIST recognition using individually trained NNs, with hidden layers ranging from 2 to 15 and learning rates from $10^{-1}$ to $10^{-8}$. Colors indicate accuracy; the vertical axis is set to a logarithmic scale.

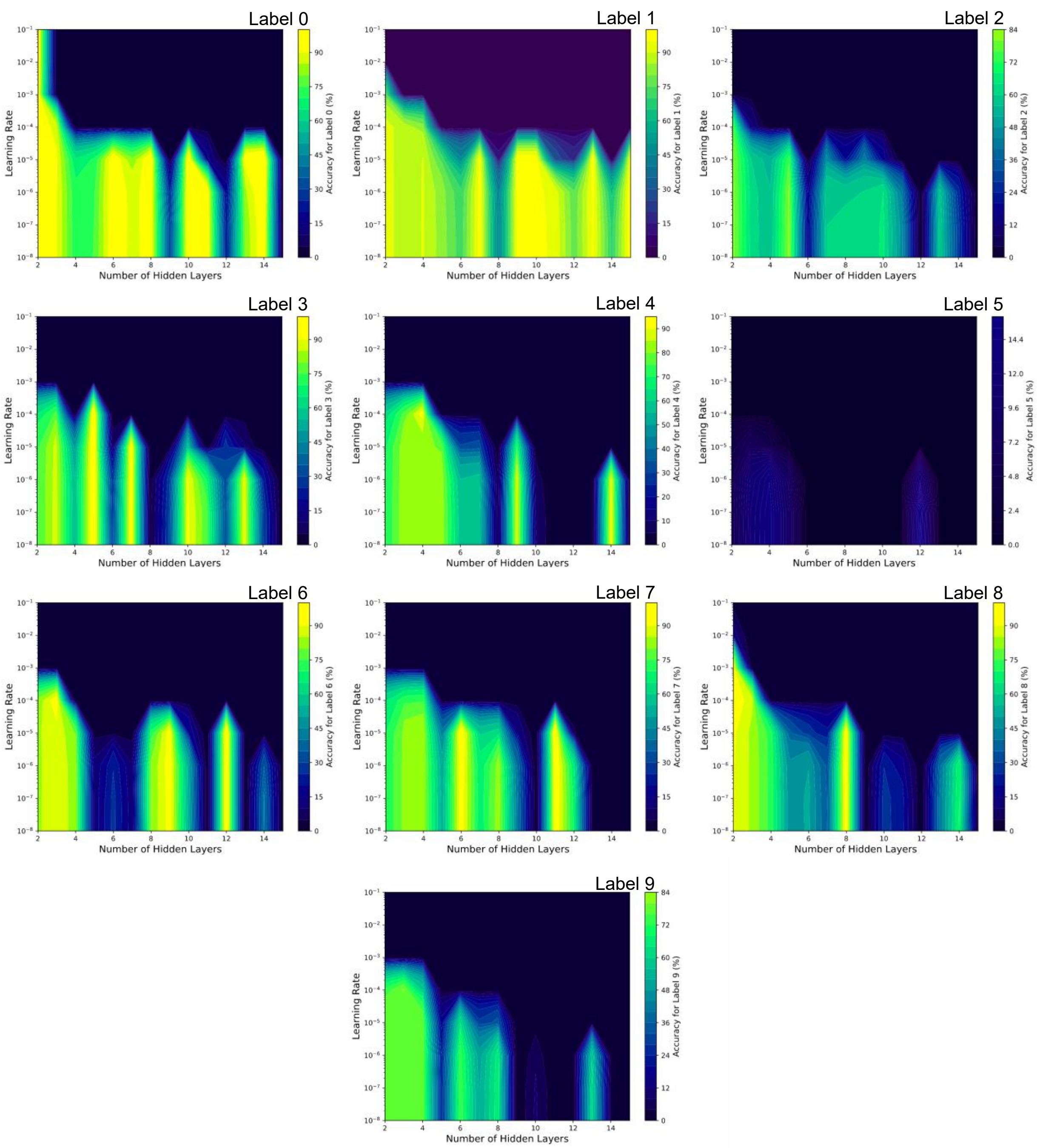

**Figure 4.** Label-specific accuracy for the MNIST recognition using individually trained NNs in Figure 3a.

Figure 3a shows that the MNIST classification program developed in this study using individually trained NNs achieved an accuracy of approximately 75% with two or three hidden layers and learning rates of $10^{-5}$ or lower. Even when the number of hidden layers was increased to 14, an accuracy of about 30% was maintained. For all hidden layer configurations, a learning rate of $10^{-5}$ or $10^{-6}$ or lower was ideal; almost no correct answers were obtained at higher learning rates. This drop-in accuracy at higher learning rates was due to the factors discussed in Section 3. The lower limit for the learning rate depends on the computer's decimal precision; if the rate is minimal, accuracy calculations cannot be performed. Figures 3b, 3c, and 3d demonstrate that significant accuracy can still be achieved when the training data are limited, as discussed in Section 4. With just one training sample (Figure 3b), an accuracy of approximately 50% was achieved. With 10 training samples (Figure 3c), the accuracy increased further, and with 100 training samples (Figure 3d), the results were almost identical to those obtained with 5420 training samples. These findings suggest that the program is capable of effectively learning from small datasets, reflecting the characteristics of Hebbian learning-based NNs. Figure 4 shows that labels 0 and 1 achieved relatively high accuracy across most hidden layer configurations, while other labels showed irregular fluctuations in accuracy depending on the number of hidden layers. However, label 5 had low accuracy under all conditions (as seen previously in Section 4 for label 5).

These results demonstrate that an NNs machine learning program can solve the MNIST classification problem with a certain level of accuracy, even without using objective functions, backpropagation structures, optimization processes, and large data set. This program, which partially mimics biological neural system and learning processes while minimizing non-biological elements, could serve as a foundation for developing biologically inspired AI or strong AI based on NNs.

## 6. Considerations on Norm-Based and Similarity-Based Cognition in NNs and Neural Systems

In this section, I discuss the cognition and decision-making mechanisms that use vector norms, as covered so far, and compare them with similarity-based cognition (from here on, I will refer to vector norms and vector similarity simply as "norm" and "similarity," respectively) to explore their application to NNs and programs that mimic biological neural systems.

Biological organisms likely lack mechanisms for directly calculating or perceiving similarity. Therefore, if one aims to create a program that mimics biological neural systems using NNs, it is likely that the conventional use of distributed representations that directly calculate similarity would be inappropriate. Furthermore, the primary mode of perception in biological neural systems should be interpreted as norm-based rather than similarity-based (4.1). For these reasons, the foundational cognition mechanism in NNs intended to mimic biological neural systems should be norm-based. In biological cognition, norm-based perception seems primarily applied to reflex actions. However, advanced cognition encompasses non-reflexive actions involving thought-based judgment beyond simple reflexive responses. Non-reflexive actions include object and character recognition and cognitive judgments based on sensory information, as observed in human cognition. These can be considered similarity-based cognition (hence why similarity-based cognition has traditionally been used in character recognition).

I hypothesize that similarity-based cognition in biological organisms does not stem from direct similarity calculations but rather from an indirect perception that combines norm-based cognition. Developing indirect similarity cognition through combinations of norm-based perceptions aligns with the goals of this study so far. Simplified norm-based cognition, as shown in the program in Section 4, enables binary choices. In Section 4, the output norm of an NNs trained on a specific label was compared with that of other NNs to identify which label the input data likely corresponded to. In Section 5, I combined individually trained NNs for all labels, allowing for recognition among ten options. This gradual shift from binary decisions in simple structures to more complex judgments through structural expansion likely leads to "indirect similarity cognition." However, Section 4 compares two NNs in parallel, so it cannot be considered the simplest form of norm-based cognition. The most basic mechanism of norm-based cognition would involve using only a single NNs to perceive "whether this is X or not." This approach aligns with reflexive actions in biological organisms, such as the sea slug's ability to close its gills based on the presence or absence of a stimulus.

There must be various methodologies for indirect similarity cognition. I used parallel NNs in this study, but this approach is not entirely biological. The brain consolidates multiple types of knowledge and memories within a single interconnected neural system rather than having separate brains for each kind of knowledge. Future development of more biologically inspired methodologies will be necessary. To achieve this, it will be essential to address simpler, biologically inspired themes rather than focusing on complex themes such as character recognition (character recognition is inherently a sophisticated function and may be too challenging as a starting point).

## 7. Conclusion

In this study, I implemented a machine learning method without error backpropagation or optimization processes using NNs with Hebbian learning and developed an MNIST handwritten character recognition program. The development was carried out in three stages. In the first stage, I applied the Hebbian learning rule to the non-machine-learning-based MNIST character recognition algorithm by Itoh (2024). This method did not yield satisfactory results and highlighted the inapplicability of conventional training procedures to Hebbian learning. In the second stage, I investigated the properties of individually trained NNs using non-distributed norm-based cognition. I demonstrated that an NNs individually trained on a specific label responded strongly to that label, providing a foundation for the practical application of Hebbian learning NNs. In the third stage, I applied the method from the second stage to create an MNIST character recognition program that uses vector norm magnitudes as the basis for judgment. This program achieved an accuracy of up to approximately 75%, demonstrating that the NNs machine learning program could perform handwritten character recognition with a certain level of accuracy without using objective functions, backpropagation structures, optimization processes, and large data set.

Additionally, based on these developments and calculation results, I considered norm-based cognition and indirect similarity cognition, which combines vector norms, as a foundation for creating programs that mimic biological neural systems. Since biological organisms lack mechanisms for directly perceiving vector similarity, developing a mechanism that uses norm-based cognition as a fundamental unit and then increases complexity to achieve indirect similarity cognition could lead to more accurate mimics of biological neural systems.